\pdfoutput=1

\documentclass[11pt]{article}

\usepackage[]{acl}

\usepackage{times}
\usepackage{latexsym}
\usepackage{graphicx}
\usepackage{amsmath}
\usepackage{url}
\usepackage{graphicx}
\usepackage{subfigure} 
\usepackage{multirow}
\usepackage{makecell}
\usepackage{colortbl}
\usepackage{amsfonts}

\usepackage{helvet}
\usepackage{courier}
\usepackage{amsmath}
\usepackage{url}
\usepackage{multirow}
\usepackage{booktabs}
\usepackage{enumitem}
\usepackage{amssymb}
\usepackage{marvosym}
\usepackage{color}
\usepackage{caption}
\usepackage{titlesec}
\usepackage{floatrow}
\usepackage{caption}
\newfloatcommand{capbtabbox}{table}[][\FBwidth]

\usepackage{colortbl}

\definecolor{shallow_grey}{RGB}{167,194,203}
\definecolor{dark_grey}{RGB}{88,98,103}
\definecolor{light_green}{RGB}{66,195,183}
\definecolor{dark_blue}{RGB}{87,133,149}

\definecolor{dark_red}{RGB}{192,1,0}
\definecolor{dark_orange}{RGB}{255,147,2}
\definecolor{dark_yellow}{RGB}{255,213,121}

\usepackage[T1]{fontenc}

\usepackage[utf8]{inputenc}
\usepackage{verbatim}
\usepackage{enumitem}

\usepackage{microtype}

\definecolor{darkgreen}{rgb}{0,0.35,0}

\title{
MORSE \protect\includegraphics[scale=.05]{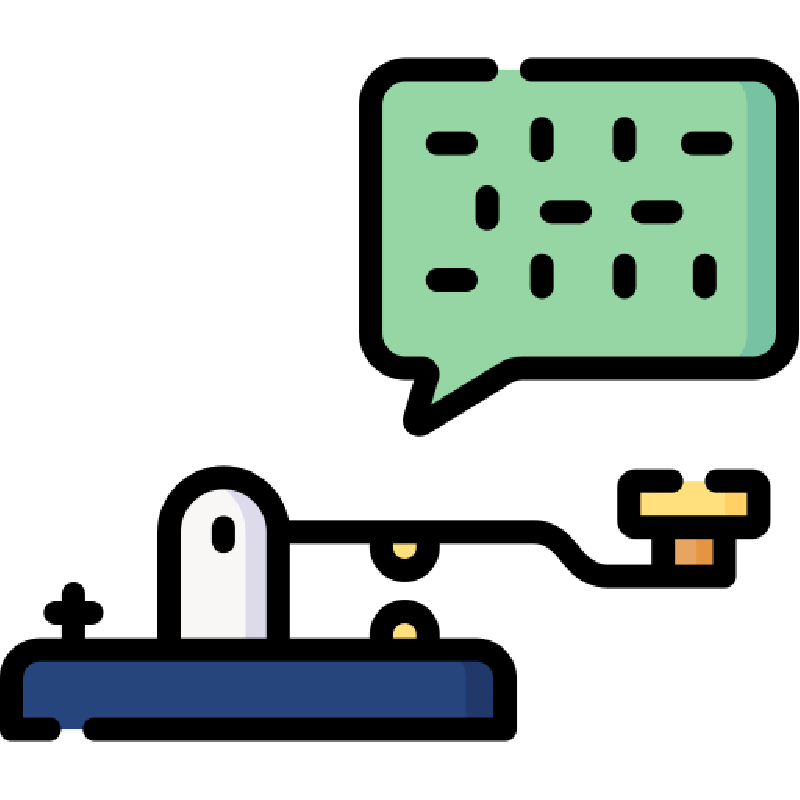} : Dynamic \underline{MO}dularized \underline{R}easoning for 
Compositional
\underline{S}tructured \underline{E}xplanation Generation}

\author{Xiyan Fu \\
  Dept. of Computational Linguistics \\
  Heidelberg University \\
  \texttt{fu@cl.uni-heidelberg.de} \\\And
  Anette Frank \\
  Dept. of Computational Linguistics \\
  Heidelberg University \\
  \texttt{frank@cl.uni-heidelberg.de} \\}

\begin{document}
\maketitle
\begin{abstract}
Despite the success of neural models in solving reasoning tasks, their compositional generalization capabilities remain unclear. In this work, we propose a new setting of the structured explanation generation task to facilitate compositional reasoning research. Previous works found that symbolic methods achieve superior compositionality by using pre-defined inference rules for iterative reasoning. But these approaches rely on brittle symbolic transfers and are restricted to well-defined tasks. Hence, we propose a \textit{dynamic modularized} reasoning model, MORSE, to improve the compositional generalization of neural models. MORSE factorizes the inference process into a combination of modules, where each module represents a functional unit. Specifically, we adopt modularized self-attention to dynamically select and route inputs to dedicated heads, which specializes them to specific functions. We conduct experiments for increasing lengths and shapes of reasoning trees on two benchmarks to test MORSE's compositional generalization abilities, and find it outperforms competitive baselines. Model ablation and deeper analyses show the effectiveness of dynamic reasoning modules and their generalization abilities.
\end{abstract}

\section{Introduction}
\begin{figure}[t]
\centering
\includegraphics[scale=0.42,trim=3 0 0 0]{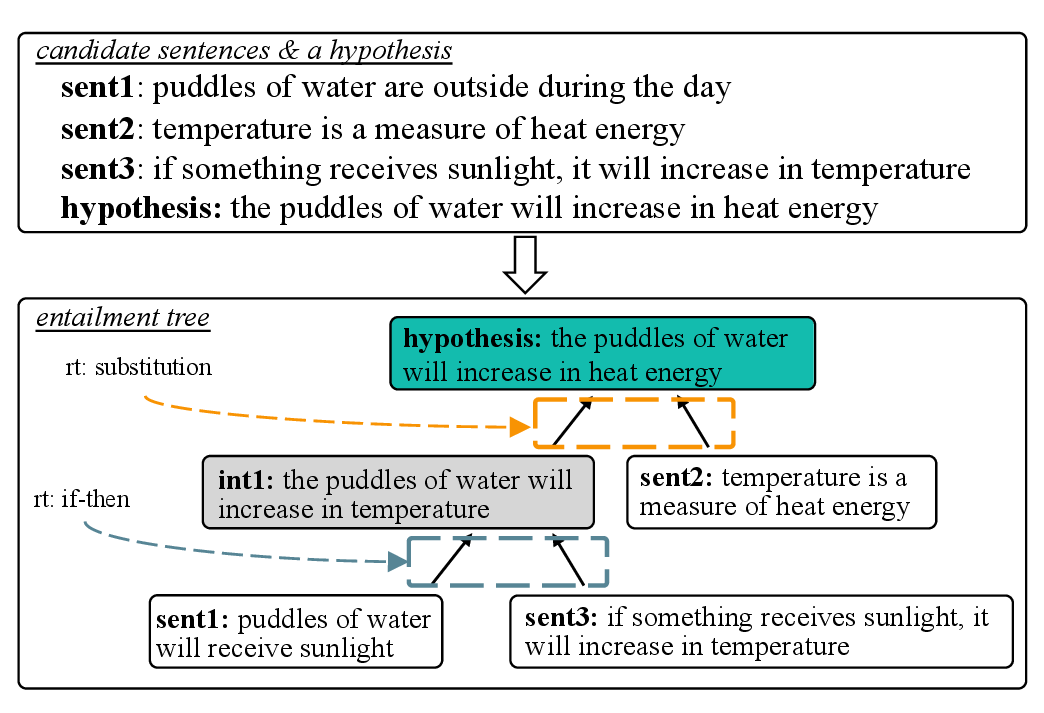}
\caption{
Structured explanation generation: generate an entailment tree including intermediate nodes (grey) for a hypothesis (green) and candidate sentences. Each reasoning step (e.g., sent1 \& sent3 $\rightarrow$ int1) is independent and from one of six reasoning types (rt).
}
\label{fig_intro}
\end{figure}

Performing reasoning, the capacity to draw conclusions from new or existing information, has remained an elusive goal of artificial intelligence research\- for decades. It yields multiple exploratory research directions, such as reading comprehension \citep{rajpurkar-etal-2018-know}, natural language inference \citep{williams-etal-2018-broad}, story generation \citep{mostafazadeh-etal-2016-corpus}, etc. Recently, neural models have shown remarkable performance on these tasks \citep{raffel2019exploring}. However, \citet{russin2020deep, mitchell2021abstraction} argued that these models lack human-like reasoning capabilities.

Humans excel in adopting \textit{compositional generalization} \citep{hupkes2020compositionality}, a capacity to combine an inventory of known constituents to predict larger compounds, during reasoning. For example, humans who understand calculation constituents, e.g., subtraction \textcolor{dark_orange}{\textit{sub(X, Y)}} and addition-subtraction mixed operation \textit{\textcolor{dark_orange}{sub(X}, \textcolor{dark_blue}{add(Y, Z)}\textcolor{dark_orange}{)}}, are able to solve a larger compound, e.g., \textit{\textcolor{dark_orange}{sub(W, sub(X}, \textcolor{dark_blue}{add(Y, Z)}\textcolor{dark_orange}{))}}.
Various benchmarks have been proposed to explore compositional generalization abilities in reasoning tasks. Some designed compositional questions \citep{johnson2017clevr, hudson2019gqa, liu-etal-2022-challenges}, others provided compositional textual inference pairs \citep{geiger-etal-2020-neural, goodwin-etal-2020-probing, yanaka-etal-2020-neural}. However, these works only focus on compositionality units manifesting on the word level and involving specific linguistic phenomena. We lack understanding of the compositionality on the \textit{sentence} level.


To fill this gap, we introduce a new setting of the structured explanation generation task (SEG) \citep{dalvi-etal-2021-explaining} to explore compositional generalization in reasoning. SEG aims to generate a multi-step entailment tree given a hypothesis and candidate sentences, showing how the hypothesis follows from the text. Fig.\ \ref{fig_intro} shows an example. Each step (e.g., sent1 \& sent3 $\rightarrow$ int1) represents a multi-premise textual inference \citep{lai-etal-2017-natural}, belonging to one of six reasoning types, such as if-then (it) and substitution (subs). We consider each reasoning type as a primitive unit. To test compositional generalization on reasoning, our new setting requires models to generalize from entailment trees with a \textit{limited} number of reasoning steps, to trees involving \textit{more} steps. For example, a model is expected to generate a larger compound (entailment tree) with more reasoning steps, e.g., \textit{\textcolor{dark_orange}{subs(subs(}\textcolor{dark_blue}{it(p$_{1}$, p$_{2}$) $\rightarrow$ h$_{1}$}\textcolor{dark_orange}{,  p$_{3}$) $\rightarrow$ h$_{2}$, p$_{4}$) $\rightarrow$ h$_{3}$}}, by combining known constituents \textcolor{dark_orange}{\textit{subs(p$_{1}$, p$_{2}$) $\rightarrow$ h}} and \textit{\textcolor{dark_orange}{subs(}\textcolor{dark_blue}{it(p$_{1}$, p$_{2}$) $\rightarrow$ h$_{1}$}\textcolor{dark_orange}{,  p$_{3}$) $\rightarrow$ h$_{2}$}}. 
Here, compositionality units, i.e., reasoning types, operate on the \textit{sentence} level and involve reasoning components.

To address this challenging task, models need to perform compositional generalization. Prior works show that symbolic-based approaches \citep{martinez-gomez-etal-2017-demand, gupta2019neural, le-etal-2022-vgnmn} which use multiple modules that each perform unique types of reasoning, endow models with strong compositionality. However, they rely on prerequisite rules and need extra training data for each pre-defined module, hence limited to well-defined tasks and symbolic systems. 
In this work, we propose a \textit{dynamically modularized} reasoning model MORSE, which aims to enhance the compositional generalization of neural models without pre-defined knowledge. 
MORSE is based on the Transformer and specializes self-attention heads to what we call \textit{dynamic modules}. We design a modularized self-attention mechanism that dynamically selects and routes inputs to dedicated modularized heads, thereby specializing them to specific functions. The dynamics embodied in MORSE through its self-assembling modules makes it applicable to multiple datasets without pre-defined knowledge and extend to novel inference types.

Our main contributions are:

\begin{enumerate}[label=\roman*), noitemsep]
    \item We propose a new setting for the structured explanation generation task, which aims to evaluate the \textit{compositional reasoning} capabilities of neural models. It requires models to generalize from entailment trees with a \textit{limited} number of inference steps to \textit{more} steps.

    \item We design a novel dynamically modularized reasoning model that specializes transformer heads to specific functions, by \textit{dynamically} selecting related inputs to dedicated heads. 
      
    \item Experiments on two benchmarks targeting generalization over proof lengths and shapes demonstrate MORSE's advanced compositional generalization abilities.
\end{enumerate}

\section{Related Work}

\paragraph{Compositional Generalization in Reasoning} Compositional generalization 
has been researched for decades \citep{fodor1988connectionism, marcus2003algebraic, lake2018generalization} and formalized in \citet{hupkes2020compositionality}. Recently, there has been a renewed interest in exploring the compositional generalization of reasoning tasks. \citet{johnson2017clevr, hudson2019gqa, bogin-etal-2021-covr, gao2022cric} proposed challenging compositional tasks in visual question answering. \citet{liu-etal-2022-challenges} designed compositional questions for question answering and found even the strongest model struggled with these challenging questions. Other works probed the compositional ability of models in natural language inference (NLI) \citep{geiger-etal-2020-neural, goodwin-etal-2020-probing, yanaka-etal-2020-neural, yanaka-etal-2021-exploring}. They focus on specific linguistic phenomena, such as quantifiers, negation, predicate replacements, etc. I.e., they only investigate compositionality in phenomena manifesting at the word level,
lacking understanding on the sentence level. To fill this gap, our work examines compositional generalization in the multi-step entailment tree generation task, which aims to compose different inference rules.

\paragraph{Neural-Symbolic and Neural Methods}
Prior works show that symbolic approaches \cite{angeli-manning-2014-naturalli, mineshima-etal-2015-higher, martinez-gomez-etal-2017-demand} that adopt pre-defined inference rules to establish the derivation through iterative reasoning, endow models with strong compositionality. However, they rely on prerequisite rules, hence limited to well-defined tasks and symbolic systems. Recently, several approaches \citep{yi2018neural, yin-etal-2018-structvae, li2020closed, jiang-etal-2021-lnn} used neural networks to map raw signals to symbolic representations and subsequently performed symbolic reasoning to make predictions. Since symbolic reasoning is brittle, novel works based on Neural Modular Networks (NMN) \cite{andreas2016neural, hu2017learning} have been introduced. These methods compose individual neural modules endowed with \textit{specialized} reasoning capabilities.
For example, \citet{jiang-bansal-2019-self} designed four novel modules based on the neural modular network to perform unique types of reasoning in an end-to-end manner. \citet{gupta2019neural} extended the number of modules to support diverse reasoning types and executed symbolic reasoning. Similarly, \citet{khot-etal-2021-text} introduced a Text Module Network for complex reasoning tasks, where each module is an existing QA system. However, all mentioned approaches require prior knowledge and rely on brittle symbolic transfer, to subsequently deploy pre-defined modules for each sub-task. Another problem is that well-designed modules require substantial extra training data. Finally, symbolic reasoning methods based on neural representations are typically driven by weak supervision, given the lack of intermediate labels. This can result in error accumulation and time-consuming learning. 

To solve these challenges, our work aims to devise a model with \textit{dynamic modules} that makes specific module functions more independent from prior knowledge, to endow models with greater flexibility when handling new tasks.

\section{Task Setup}
\paragraph{Task Definition} The Structured Explanation Gen\-era\-tion (SEG) task \citep{dalvi-etal-2021-explaining} requires a system to generate reasoning chains as structured explanations of how presented evidence leads to a conclusion. Inputs consist of two parts: 1) hypothesis $H$ is a declarative statement; 2) a set $S$ of candidate sentences express relevant knowledge needed to infer the hypothesis. Systems are tasked to output valid entailment trees with intermediate conclusions not contained in $S$ (Fig.\ \ref{fig_intro}). Leaves $sent_{*}$ are sentences from the candidate set $S$, intermediate nodes $int_{*}$ are interim conclusions and the tree's root node is the hypothesis $H$. We expect each intermediate node $int_{*}$ to be entailed by its children, e.g., $sent1$, $sent4$ induce the interim conclusion $int1$. The inferred intermediate conclusions provide the basis for further reasoning steps.

\paragraph{Compositional Generalization Test}
To examine compositional generalization capabilities in reasoning, we rearrange the partitions of our benchmark datasets focusing on two properties \citep{hupkes2020compositionality}: \textit{productivity} and \textit{systematicity}.

\begin{figure}[t]
\includegraphics[width=2.9in]{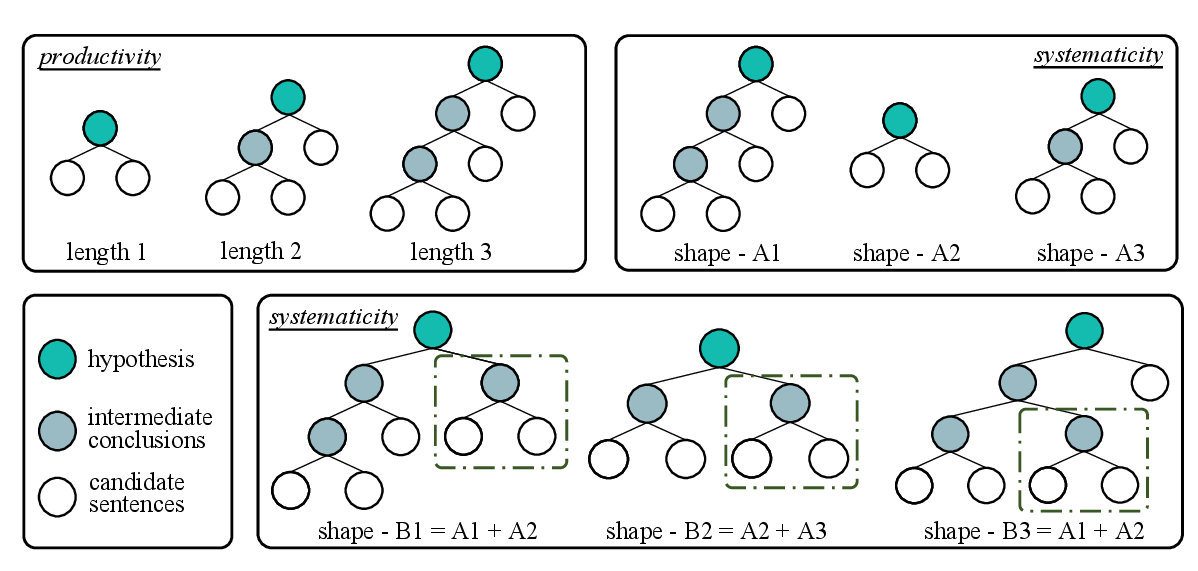}
\caption{Entailment trees including different lengths and shapes for compositional generalization test.}
\label{fig_shape}
\end{figure}

\textbf{Productivity-Length} Productivity refers to the capacity to grasp an indefinite number of primitives, which evaluates systems on longer sequences than they have been trained on.
Hence, we rearrange the data by proof length, i.e., the number of intermediate nodes in each tree (including the hypothesis node). Data partitions are constructed as: i) primitive proof instances of length one or two; ii) compositional proof instances of length three\footnote{We only test length three here, given the significant challenge shown by experiments. However, our setting is a living benchmark, which can be easily extended by future research.}. Fig.\ \ref{fig_shape} shows examples.

\textbf{Systematicity-Shape} Systematicity examines if a model can recombine constituents that have not been seen together during training.
Hence, we rearrange the dataset by proof shapes. Fig.\ \ref{fig_shape} shows the entailment tree shapes in the two partitions: \textbf{Shape-A*} contains proofs of primitive shapes: A1-A3; \textbf{Shape-B*} proof shapes result from \textit{compositions} of shape-A* shapes: B1=A1+A2, B2=A2+A3, B3=A1+A2. 
We create partitions for: i) primitive proof instances of Shape-A*; ii) compositional proof instances of Shape-B*.

\section{MORSE: Dynamic Modularized Reasoning Model}
\begin{figure*}[t]
\centering
\includegraphics[scale=0.55,trim=3 0 0 0]{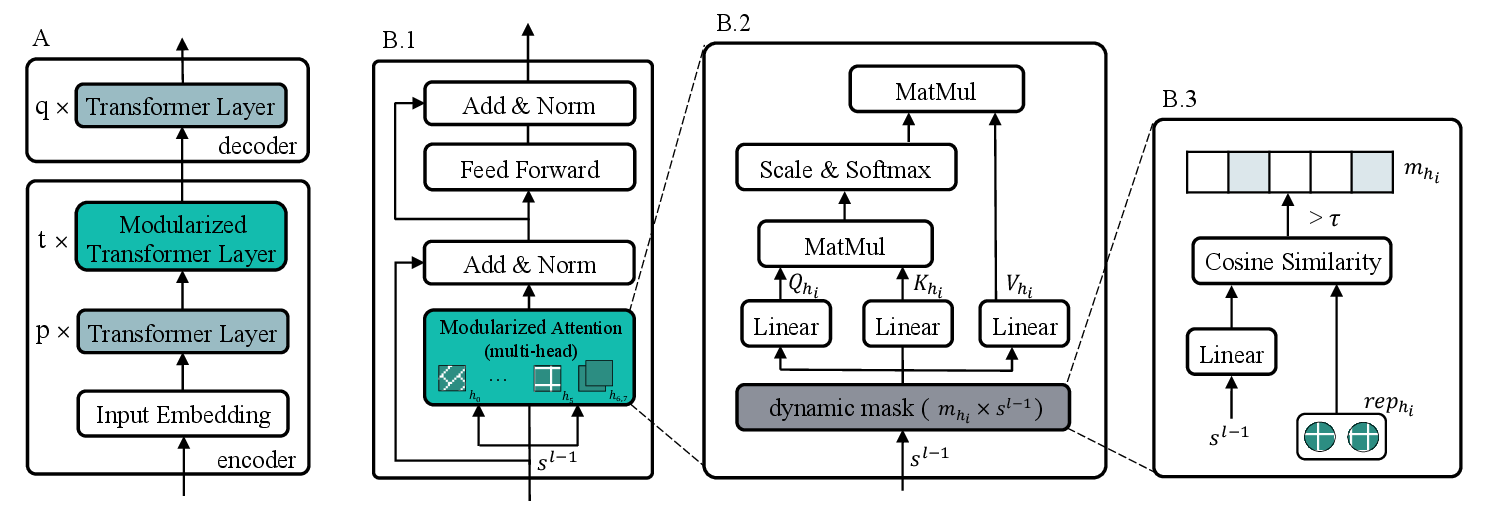}
\caption{(A) MORSE for entailment tree generation. (B) A series of detailed illustrations of the Modularized Transformer layer. (B.1) Our novel \textit{modularized} multi-head self-\-attention block. Each head may serve as a module, executing a specific function. (B.2) Computations for a single attention head with dynamic mask $m_{h_{i}}$. Self-attention is extended with a dynamic mask to filter out irrelevant input for a module. (B.3) Constructing dynamic mask $m_{h_{i}}$ using head function representation $rep_{h_{i}}$ and input hidden states.}
\label{fig_model}
\end{figure*}

We introduce our Dynamic Modularized Reasoning Model MORSE that generates compositional structured explanations. MORSE contains: i) an encoder consisting of original and modularized transformer blocks to perform reasoning; ii) a decoder using original transformer blocks to generate the entailment tree structure. 
See the overview in Fig.~\ref{fig_model}.\par

\subsection{Module-enhanced encoder}
We concatenate candidate sentences $S$ and the hypothesis to an input sequence. For each sentence in $S$, we add a prefix $sent*$ following \citet{dalvi-etal-2021-explaining}. Thus the example in Fig.\ref{fig_intro} would be represented as a sequence of length $n$: `sent1: puddles of water are outside during the day; sent2: temperature ...; ...; hypothesis: the puddles of water will increase in heat energy'. For each token $x_{i}$, we adopt an embedding layer to generate its representation $e_{i}$, i.e., a summation of token embedding, position embedding and segment embedding. An encoder subsequently encodes input representations.

Fig.\ref{fig_model}.A shows that MORSE's encoder consists of \textit{Transformer} blocks for lower layers and \textit{Modularized Transformer} blocks for higher layers: i) Transformer blocks allow the model to focus on the representation of words themselves \citep{raganato-tiedemann-2018-analysis, jawahar-etal-2019-bert}; ii) Modularized Transformer blocks perform modularized reasoning, where each module is encouraged to take on a different inference function. 

\paragraph{Transformer}
Each Transformer block consists of two sub-layers: a multi-head attention layer and a fully connected feed-forward network. Each sub-layer is followed by layer normalization \citep{ba2016layer} and a residual connection \citep{he2016deep}. In the multi-head attention sub-layer, sequential inputs are projected to different representation sub-spaces (different heads) in parallel; the layer then performs self-attention \citep{vaswani2017attention} in each head. Output values from these heads are concatenated and once again projected, resulting in the final values.

In MORSE, we adopt $p$ Transformer blocks in lower layers, aiming to capture the representation of words 
in their syntactic context. Given token embeddings $e_{1}$, ..., $e_{n}$ of a sequential input of length $n$, we use $p$ Transformer blocks to encode them and generate corresponding hidden states $s_{1}^{p}$, ..., $s_{n}^{p}$.

\paragraph{Modularized Transformer}
We construct a Mo\-du\-la\-ri\-zed Transformer block based on the Transformer. The difference is that we factorize the encoding process, by modularizing the Transformer so that each module can be tailored to a specific function. We implement this design by using Transformer `heads'. The process of \textit{modularization} is illustrated in Fig.\ \ref{fig_model} B.1: the modularized Transformer block contains a modularized attention layer, which consists of multiple specialized heads $h_i$. E.g., $h_{0}$ to $h_{5}$ are modularized heads that may express different inference functions. The remaining heads $h_{6,7}$ work as usual, offering space to model general knowledge not covered by the modularized heads. With such modularization, we expect that each module 
will specialize for specific responsibilities,
further endowing MORSE with more flexibility to perform different inference functions during reasoning.

To allow a modularized head $h_i$ to specialize for specific functions, we construct dynamic masks $m_{i} \in [0,1]^{n}$ to select sequential inputs of similar kinds to pass through $h_i$. Specifically, we define several vectors of trainable parameters for each module as a latent representation of the modules' function, e.g., $rep_{h_{i}} \in \mathbb{R}^{d}$ for $h_{i}$. Simultaneously, we adopt a linear projection on candidate input hidden states $s_{1}$, ..., $s_{n}$ to derive their functional representations $f_{1}$, ..., $f_{n} \in \mathbb{R}^{d}$. Then, we use cosine similarity $cos$ over the input's functional representations $f_{j}$ and the head's representation $rep_{h_{i}}$ to calculate a matching coefficient. If it exceeds a threshold $\tau$, MORSE is able to decide if an input word $x_{j}$ is allowed to join the module $h_{i}$. The calculation is shown below:
\begin{equation}
m_{i}^{j} = 
\left\{
\begin{array}{ll}
exp^{1-cos(rep_{h_{i}}, {f}_{j})}, & cos(rep_{h_{i}}, {f}_{j}) > \tau \\
0, & else
\end{array}
\right.
\end{equation}
where the threshold $\tau$ is a fixed hyper-parameter. Here, we keep the gradient for selected words and ignore the remains. We guarantee the gradient computation for selected word representations by approaching 1 with $exp^{1-cos(*)}$. In this way, we can generate masks $m_{i}$ for each module $h_{i}$ dynamically, given sequential inputs and different module objectives.

We further adopt the generated mask $m_{i}$ for a module $h_{i}$ in Modularized Self-Attention to filter out unrelated inputs. Fig.\ \ref{fig_model} B.2 shows the process: we multiply the mask $m_{i}$ with input hidden states from the previous layer $s^{l-1}$, where hidden states of unrelated words are set to zero. Then, we generate the query $Q_{h_{i}}$, key $K_{h_{i}}$, and value $V_{h_{i}}$ matrices for self-attention by different linear projections based on filtered inputs:
\begin{equation}
\begin{aligned}
    Q_{i}, K_{i}, V_{i} &= \widetilde{s}^{l-1}W_{i}^{Q}, \widetilde{s}^{l-1}W_{i}^{K}, \widetilde{s}^{l-1}W_{i}^{V} \\
    \widetilde{s}^{l-1} &= m_{h_{i}} \times s^{l-1}
\end{aligned}
\end{equation}
where $W_{i}^{Q}, W_{i}^{K}, W_{i}^{V}$ $\in \mathbb{R}$ $^{d \times d/k}$ are training parameters, $d$ is the hidden state dimension and $k$ is the number of heads. We then adopt scaled dot-product attention to perform self-attention:
\begin{equation}
    a_{i} = softmax(\frac{Q_{i}K_{i}^{T}}{\sqrt{d_{k}}})V_{i}
\end{equation}

\noindent
We adopt $t$ Modularized Transformer blocks in deep layers, aiming to perform modularized reasoning. Given input hidden states $s_{1}^{p}, ..., s_{n}^{p}$ from lower Transformer blocks, the Modularized blocks generate modularized hidden states $s_{1}^{t}, ..., s_{n}^{t}$.

\subsection{Decoder and training}
We use a decoder consisting of Transformer blocks to generate the entailment tree structure and intermediate conclusions. The tree is linearized from leaves to the root. For example, the tree in Fig.\ \ref{fig_intro} is represented as ``sent1 \& sent4 $\rightarrow$ int1: \textit{puddles of water will receive sunlight}; sent3 \& int1 $\rightarrow$ int2: \textit{the puddles of water will increase in temperature}; sent2 \& int2 $\rightarrow$ hypothesis'. The output sequence generation process is defined as:
\begin{equation}
\begin{aligned}
    s^{l} = block(s^{l-1}, enc\_state), \quad l \epsilon [1, q] \\
    p(y_{k}|y_{<k}) = softmax(s_{k}^{N}W^T)
\end{aligned}
\end{equation}
where $s^{l}$ is the $l_{th}$ layer computed through Transformer blocks, $W^T$ is the training parameter and $k$ is the decoding step number. We deploy supervised learning with ground truth by minimizing the objective in (5), where M is the maximum length of the generated proof, $H$ and $S$ are hypothesis and candidate sentences, respectively.

\begin{equation}
    L = - \sum_{k=1}^{M} log p(y_{k}|y_{<k}, H, S)
\end{equation}

\section{Experiments Setup}

\subsection{Datasets}
We evaluate our model MORSE on two benchmarks that use an unconstrained number of possible inference rule types to generate multi-step tree-structured explanations: EntailmentBank (EntB) and DBpedia (DBP).

\textbf{EntailmentBank} \citep{dalvi-etal-2021-explaining} contains multiple-choice questions and candidate sentences from the grad-school level science datasets ARC \citep{clark2018think} and WorldTree \citep{jansen-etal-2018-worldtree,xie-etal-2020-worldtree}. The 1,840 entailment trees each show how a hypothesis is entailed from a small number of relevant sentences. Each step in the tree represents an entailment, i.e., the conclusion expressed in each intermediate node follows from the content of its immediate children. The individual entailment steps instantiate six common reasoning types (see \ref{app:EntailmentBank} for more details). EntB contains three tasks. We mainly focus on Task1, which uses only correct inputs in $S$, as we focus on generalization testing.

\textbf{DBpedia} \citep{saeed-etal-2021-rulebert} is a synthetic data\-set that was re-generated from the \textbf{RuleBert} data\-set\footnote{\url{https://github.com/MhmdSaiid/RuleBert}}. We extracted six distinct logic rules mined from the DBpedia knowledge graph and instantiated examples with a varying number of variables following `Chaining of Rule Execution' in RuleBert (see \ref{app:DBpedia} for more details). The reasoning chain provides a structured explanation: each intermediate node is a conclusion inferred from immediate children using a logic inference rule.

\begin{table*}[t]
\centering
\resizebox{1\columnwidth}{!}{
\begin{tabular}{@{}lcccccccccccc@{}} \hline
& \multicolumn{6}{c}{EntailmentBank-Length (EntB-L)} & \multicolumn{6}{c}{DBpedia-Length (DBP-L)} \\ \cmidrule(r){2-7} \cmidrule(r){8-13} 
\multirow{3}{*}{Models} & \multicolumn{2}{c}{Leaves} & \multicolumn{2}{c}{Steps} & \multicolumn{2}{c}{Intermediates} & \multicolumn{2}{c}{Leaves} & \multicolumn{2}{c}{Steps} & \multicolumn{2}{c}{Intermediates}\\ \cmidrule(r){2-3} \cmidrule(r){4-5} \cmidrule(r){6-7} \cmidrule(r){8-9} \cmidrule(r){10-11} \cmidrule(r){12-13} 
 &F1  &AllCorrect &F1  &AllCorrect  &F1  &AllCorrect &F1  &AllCorrect &F1  &AllCorrect  &F1  &AllCorrect   \\ \cmidrule{1-13}
ProofWriter-It.  &91.86(0.08) &84.55(0.78) &35.97(2.37) &18.81(2.76) &42.93(1.23) &11.88(2.14) &90.66(0.18) &93.09(0.72) &76.49(0.86) &75.44(1.04) &85.92(1.92) &76.73(2.24) \\ 
PROVER  &- &- &39.27(2.65) &24.75(3.24) &- &- &- &- &79.88(0.98) &76.98(1.37) &- &- \\
EntWriter (T5-Small) & 99.78(0.12) & 98.02(1.06) & 40.59(2.97)  & 29.70(2.92)  & 48.24(1.12)   & 22.77(2.25)  &99.92(0.15) &99.49(0.67) &82.01(1.21) &79.28(1.52) &87.05(2.23) &78.26(2.37)      \\ 
EntWriter (T5-Large) &99.78(0.11) &98.02(0.99) &52.80(3.35) &40.92(3.18) &56.62(1.06) &36.63(2.40) &99.36(0.13) &95.52(0.91) &82.49(1.09) &80.11(1.43) &88.98(2.16) &83.89(2.15) \\ \cmidrule{1-13}
MORSE (T5-Small) &99.89(0.08) & 99.01(0.62) & 44.22(2.14) & 32.67(2.32) & 50.66(0.68) & 25.74(1.92) &99.96(0.27) &99.74(0.84) &82.27(0.16) &80.31(0.18) &87.72(1.82) &79.80(1.87) \\ 
MORSE (T5-Large) &\textbf{99.82}(0.06) &\textbf{98.68}(0.57) &\textbf{53.31}(2.26) &\textbf{42.57}(2.62) &\textbf{57.78}(0.81) &\textbf{37.29}(2.06) &\textbf{99.53}(0.11) &\textbf{96.68}(0.73) &\textbf{86.79}(0.12) &\textbf{83.76}(0.18) &\textbf{92.62}(1.70) &\textbf{86.70}(1.97) \\\cmidrule{1-13}\cmidrule{1-13} \hline
EntWriter-0-shot (T5-L)  &97.06(0.66) &85.73(1.61) &18.44(1.18) &- &24.21(2.22) &- &90.09(0.42) &29.27(0.2) &16.94(1.68) &- &32.43(0.50) &-      \\
MORSE-0-shot (T5-L) &\textbf{97.89}(0.74) &\textbf{86.83}(1.52) &\textbf{19.14}(0.89) &- &\textbf{25.42}(1.49) &- &\textbf{89.82}(0.32) &\textbf{30.05}(0.90) &\textbf{18.41}(1.09) &- &\textbf{33.45}(0.22) &- \\ \cmidrule{1-13}
\end{tabular}}
\caption{Results on EntailmentBank-L(ength) and DBpedia-L(ength) for compositional generalization evaluation. All modules are evaluated with 3 rounds, we show mean accuracy (std).}
\label{table_main_rst}
\end{table*}

To evaluate the compositional generalization capabilities of modularized reasoning with MORSE, we rearrange the partitions of the above benchmarks to focus on \textit{productivity} and \textit{systematicity} following \S 3 (see \ref{app:EntailmentBank_construct} for details). We construct i) EntB(ank)-L(ength) and DBP-L(ength) based on proof length; and ii) EntB-Sh(ape) based on proof shape. Since DBpedia does not contain more complex tree shapes, it is ignored in the shape test. For data statistics of the created splits for length and shape testing, see
Appendix \ref{app:data_statics}.

For reference, the results obtained by our model MORSE on
the original structured explanation generation task are reported
in Appendix \ref{app:original_rst}.

\subsection{Model Details}
For a fair comparison, we built MORSE on T5-Small/-Large following  \citep{raffel2019exploring} \footnote{Experiments on more powerful backbones are provided in Appendix \ref{app:powerful_back}.}, with six or twelve layers. 
For each version, we select lower layers (30\%) as the original Transformer blocks to derive hidden representations of the words themselves, i.e., two or four layers.

\textit{Hyperparameters.} 
We use the T5 checkpoints from Huggingface \citep{wolf-etal-2020-transformers}. For initialization, we treat all layers as plain transformer layers. We optimize our model using Adam Op\-ti\-mizer \citep{kingma2014adam} with learning rate 1e-4 and batch size 4. In inference, we adopt beam search decoding with beam size 3 for all models and baselines. We set the threshold $\tau$ for dynamic mask construction to 0.1 (details in Appendix \ref{app:parameters}). We use 20\% of training or fine-tuning datasets for validation. All models are evaluated with 3 rounds.

\subsection{Baselines}
We compare MORSE to three representative baselines of generating structural explanations: \par

\textbf{EntailmentWriter} \citep{dalvi-etal-2021-explaining} is a T5-based seq-to-seq generative model that generates a structured explanation (tree) directly. It provides baseline results on EntailmentBank for generating en\-tail\-ment trees for answers to science questions. \par

\textbf{PROVER} \citep{saha-etal-2020-prover} jointly answers binary questions over rule-bases and generates the corresponding proofs. The model learns to predict edges corresponding to proof graphs using multiple global constraints. Since PROVER focuses on edge prediction, we only evaluate the tree structure.\par

\textbf{ProofWriter-Iterative} \citep{tafjord-etal-2021-proofwriter} 
iteratively generates 1-step conclusions and proofs, adds intermediate conclusions to the context and assembles a final proof chain from 1-step fragments.
\par

\subsection{Automatic Evaluation Metrics} 
We adopt the evaluation metrics proposed by \citet{dalvi-etal-2021-explaining} for the structured explanation generation task. Evaluation is addressed in two steps: \par

1) \textbf{Alignment} Exact matching between a predicted entailment tree $T_{pred}$ and a human-labelled entailment tree $T_{gold}$ ignores the different organizations among tree nodes and leads to an inaccurate evaluation score. To admit semantic variation, all $T_{pred}$ nodes are (greedily) aligned to nodes in $T_{gold}$ using the sent* labels for leaf nodes, and use Jaccard similarity calculation for intermediate nodes.

2) \textbf{Score} Once aligned, three metrics measure the degree of similarity of $T_{pred}$ and $T_{gold}$: (a) \textit{Leaves} evaluates if the generated tree selects the correct leaf sentences from the candidate set $S$. (b) \textit{Steps} assesses if the individual entailment steps in the tree are structurally correct. This is the case if for a pair of aligned intermediate nodes, both children have identical labels (sent* or int*) in $T_{pred}$ and $T_{gold}$. (c) \textit{Intermediates} judges if all generated intermediate conclusions are correct. BLEURT \citep{sellam-etal-2020-bleurt} with the threshold 0.28 is applied for intermediate conclusion evaluation. For each metric, we compute an F1 score, and an `AllCorrect' score for exact tree matching (F1=1).

\section{Results}
Zero-shot compositional generalization is highly non-trivial due to the long generated texts of the compositional samples.\footnote{The difficulty is primarily due to the decoder trained by maximum likelihood, which relies heavily on the distributional characteristics of the dataset and assigns low probabilities to unseen combinations in test \citep{holtzman2020curious}} We therefore consider a flexible
learning scenario, where a handful of compositional examples are included in the training, following \citet{bogin-etal-2021-covr, yin-etal-2021-compositional}.

\subsection{Overall Results}
\textbf{Results on Length Composition} Table \ref{table_main_rst} displays the results of MORSE using the small vs.\ large T5 model as backbone, on the EntB-L and DBP-L datasets. Note that PROVER \citep{saha-etal-2020-prover}, EntailmentWriter (EntWriter) \citep{dalvi-etal-2021-explaining} and MORSE generate the complete proof chain from the input candidate set in one go, while ProofWriter-Iterative (PW-Iterative) \citep{tafjord-etal-2021-proofwriter} generates one-step implications iteratively. We find that on both datasets, and for both T5 model sizes, MORSE achieves superior results compared to all baselines, especially on `Steps' (structural correctness) and `Intermediates' (intermediate conclusions). `Leaves' is not a challenge in our Task1 setup, but even here, MORSE outperforms, being able to integrate almost all inputs. The comparison with the most competitive baseline model EntWriter, in equivalent T5 model sizes, still shows superior performance of MORSE with both model sizes. We conclude that the advance of MORSE is not restricted to small models, but persists with models hosting richer knowledge. Compared to DBP-L, the advance of MORSE over its competitors is stronger on EntB-L (e.g., +2.97 vs.\ +1.03 for `Steps Acc'). This is explained by the synthetic (template-based) nature of the DBP-L dataset, which shows little linguistic variety. \par

\begin{figure}[t]
    \centering
    \includegraphics[width=2in]{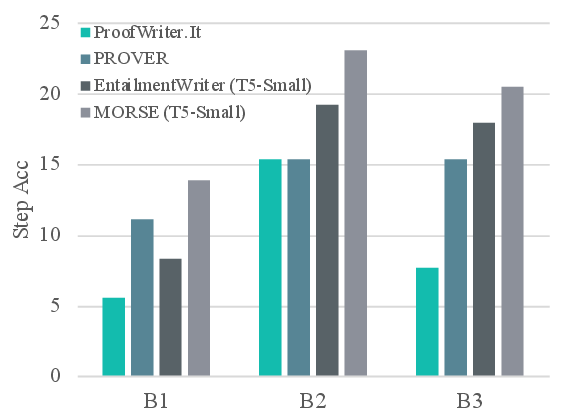}
    \caption{Results on EntB-Sh, testing for compositional generalization, i.e., systematicity.}
    \label{fig_shape_rst}
\end{figure}

To provide a comprehensive evaluation of the proposed new setting for future research, we further challenge MORSE by exposing it to a \textit{zero-shot test} for length composition. Here, models trained only for trees up to depth two will directly receive inputs for proof trees of length three. We mainly compare with the most competitive baseline, EntWriter.
In this evaluation, we ignore the `AllCorrect' scores for `Steps' and `Intermediate' outputs, given the difficulty of these generation tasks in low training regimes.
The last two lines in Table \ref{table_main_rst} show the results. 
MORSE achieves superior performance (at least +1 point improvement for zero-shot) for most evaluation categories, or else comparable results (F1 for  `Leaves'). We conclude that our model MORSE outperforms the baseline in both zero-shot and fine-tuning scenarios. 

\textbf{Results on Shape Composition}
Fig.\ \ref{fig_shape_rst} displays the results for generalization testing on shapes.\footnote{Having seen linear behaviour of different model sizes in Table \ref{table_main_rst}, we further on use T5-Small versions of MORSE and EntWriter, unless we explicitly say otherwise.} MORSE clearly surpasses the step accuracy of all baselines for all tested shape configurations. We note that shape B1 is most difficult for all systems. Entailment trees are linearized in bottom-up order. While the compositions in Shape B2 and B3 happen at the lowest tree level, B1 composition happens at a higher tree level, combining trees of unequal depths. We hypothesize that combining trees of unequal lengths at higher levels makes the task more challenging compared to lower levels,
given that composition at higher levels requires a more precise representation of previous reasoning steps.

\begin{table}[t]
\centering
\small
\resizebox{\columnwidth}{!}{
\begin{tabular}{@{}lclcl@{}} \hline
\multirow{2}{*}{Models} & \multicolumn{2}{c}{Steps} & \multicolumn{2}{c}{Intermediates} \\ \cmidrule(r){2-3} \cmidrule(r){4-5} 
&F1  &Acc  &F1  &Acc      \\ \hline
MORSE {\small (T5-Small)} & 44.22 & 32.67 & 50.66 & 25.74 \\ \hline
freeze rep\_embed  &43.57  &31.68 (-0.99)  & 50.66  &25.74 (-0)     \\
+ module &41.58 &29.70 (-2.97)   &49.13 &23.76 (-1.98)   \\ 
+ masking  &38.28 &25.74 (-6.93)   &46.62   &20.79 (-4.95)   \\ \hline
\end{tabular}
}
\caption{Ablation of MORSE components, freeze: \textbf{rep\_embed}: the representation of module $rep_{i}$; \textbf{module}: parameters in specialized module;  \textbf{masking}: dynamic mask in Fig.\ \ref{fig_model}. d. Brackets: decrease in accuracy.}
\label{table_ablation}
\end{table}

\subsection{Analysis of Modularization}
\textbf{Ablation Study} To gain more insight into the impact of specific components of MORSE on generalization, we run an ablation study on EntB-L during \textit{fine-tuning} phase. We first freeze the module representation $rep_{i}$ (\textit{-rep\_embed}) and further parameters in each specialized module (\textit{-module}). By freezing parameters, we preserve the module functions, but expect a comparative performance by re-using learned functions. Finally, we freeze the parameters of the dynamic mask process \textit{-masking} (cf.\ Fig.\ \ref{fig_model}.B.3), 
which affects the dynamic mask of inputs to different modules. Results in Table \ref{table_ablation} indicate that the first two settings do not affect results much, which suggests that each module's head has learned how to compose primitives. But \textit{-mask} incurs large drops, which indicates that masking is significant for the model to adapt to novel configurations. We hypothesize that for generalizing to longer proofs, mask generation helps to connect existing modules.

\begin{figure}[t]
\centering
\includegraphics[scale=0.43,trim=3 0 0 0]{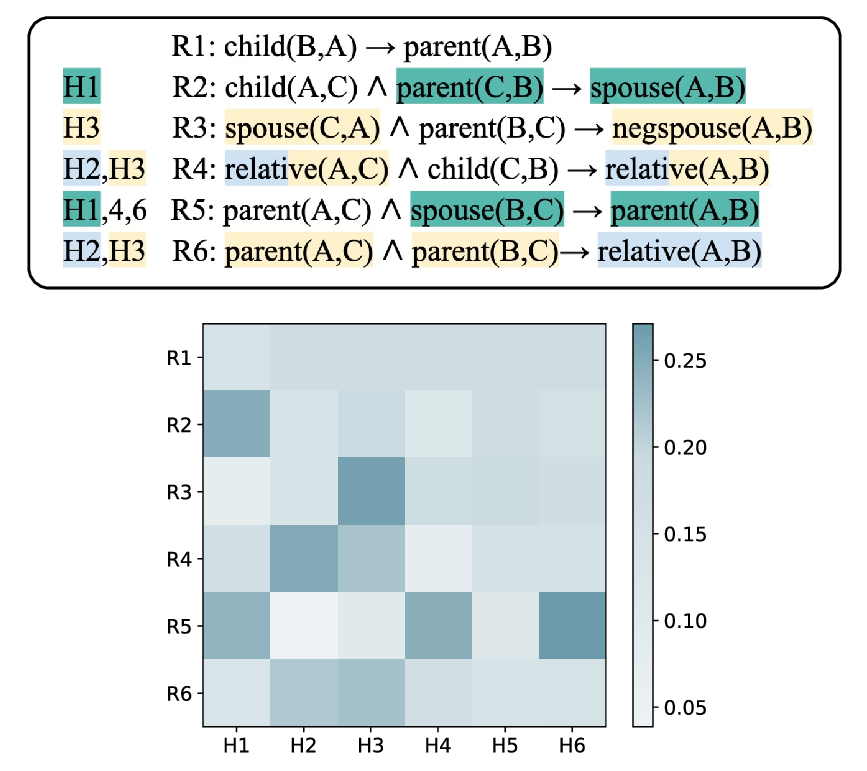}
\caption{Correlations between 
reasoning rules R1-6 and module heads H1-6. See App.\ \ref{app:corr_ana} for deeper analysis.}
\label{fig_case}
\end{figure}

\textbf{Correlation Analysis} To further explore the effects of modularization in MORSE, we conduct an experiment on DBP-L by \textit{masking individual heads} only in testing. We select samples that: i) contain three reasoning steps, ii) made correct predictions for the first two reasoning steps, but iii) predict the 3$^{rd}$ step incorrectly in case a certain head is removed (see \ref{app:module_case} for details). This ensures that the reasoning rule for the 3$^{rd}$ step is affected by a specific removed head. 
We count samples that are affected by removing head $j$ for each rule $R_{i}$, denoted as $n^{R_{i}}_{j}$. In case a model has $T$ heads, we normalize affected sample counts of $R_{i}$ across all heads, i.e., $n^{R_{i}}_{j} / \sum_{j=1}^{T}n^{R_{i}}_{j}$. This allows us to align heads and rules as shown in Fig.\ \ref{fig_case}. 

The heatmap shows the correlations between rules and heads, where R2-H1, R3-H3, R4-H2/H3, R5-H1/H4/H6 and R6-H2/H3 stand out. In the upper part of Fig.\ \ref{fig_case} we list all inference rules from DBP-L, aligned with the heads they are strongly correlated with, according to the heatmap. 
We find that heads are correlated with some rules roughly: 1) H4 and H6 are quite similar, and both prefer R5. 2) H1 prefers R2, but is distracted by R5. This is likely because R2 and R5 are similar by changing `parent' to `child' between A and C. 3) H2 prefers R4 and R6. 4) H3 prefers R3, but is distracted by R4 and R6. A plausible reason could be configurations of R3, R4 and R6 are similar as they share similar predicates (`spouse' in R3, `relative' in R4 and `parent' in R6). 

\begin{table}[tbp]
\resizebox{\columnwidth}{!}{
\begin{tabular}{@{}lcccccc@{}} \hline
Head &H1  &H2 &H3  &H4 &H5  &H6 \\ \hline
non-zero (\%) &21.46  &22.14 &21.11 &33.13 &41.31 &21.18   \\ \hline
\end{tabular}}
\caption{Non-zero values in masks for each module (\%).}
\label{table_sparse}
\end{table}

\begin{table}[t]
\centering
\small
\resizebox{\columnwidth}{!}{
\begin{tabular}{@{}lcccc@{}} \hline
\multirow{2}{*}{Models} & \multicolumn{2}{c}{Steps} & \multicolumn{2}{c}{Intermediates} \\ \cmidrule(r){2-3} \cmidrule(r){4-5} 
&F1  &Acc  &F1  &Acc      \\ \hline
w modularized\_mask  &44.22 &32.67 &50.66 &25.74    \\  \hline
w random\_mask (20\%) &30.36 &15.84  &42.62 &13.86   \\ 
w random\_mask (50\%) &36.63 &20.79  &45.45 &18.81   \\ \hline
\end{tabular}
}
\caption{Effects of different mask strategies. (*\%) indicates *\% percentage of non-zero value in a mask.}
\label{table_mask_effect}
\end{table}

\subsection{Analysis of Dynamic Masking Mechanism}

\textbf{Mask Sparsity} MORSE deploys masks to modularize a network dynamically. This allows each module to specialize for a specific function while selecting corresponding inputs. To gain more insight into the role of dynamic masking, we analyse masks used in length generalization testing on EntB-L. We count the number of masks with non-zero values for each module. Table \ref{table_sparse} shows that the percentage of \textit{non-zero values} for heads H1-6 is relatively low, indicating that dynamic masks are effective for filtering out potentially irrelevant inputs. We also note higher percentages for some modules (e.g., H4, H5). Different reasoning types require disparate inputs that may account for this.


\textbf{Mask Effects}
We apply different masking strategies to test if the observed performance improvements arise from modularized masks -- as opposed to na\"ive ones. We construct a \textit{random\_mask} model variant with 20 and 50\% non-zero values, respectively. These proportions are similar to what we find in MORSE (Tab.\ \ref{table_sparse}). We apply random masks in length composition testing on the EntB-L dataset. Table \ref{table_mask_effect} shows that compared to dynamic routing in MORSE, random masking incurs a severe performance drop. We conclude that i) unselective masking risks shielding important information from heads, and that ii) dynamic routing cannot be considered as a simple dropout mechanism.



\section{Conclusion}
We present a new setup for explanation generation to facilitate compositional reasoning research. Inspired by highly compositional symbolic systems, we propose a novel modularized reasoning model MORSE that factorizes reasoning processes into a combination of \textit{dynamically} specializing modules. MORSE outperforms competitive baselines on two benchmarks. A promising future direction is to learn 
how to initialize more modules on demand.

\section{Limitations}
The dynamic modularized reasoning model MORSE in its current state is limited by \textit{Module Number}. MORSE assumes a pre-defined number of modules for reasoning in various scenarios. The number of modules in MORSE interacts with the ability of the model when modularizing a given number of potential logic rules in a dataset or task. A given available number of functional units can simplify the reasoning process, enabling the model to focus on module re-use similar to how a symbolic system does, instead of distracting from confirming module function granularity.


\bibliography{anthology,custom}
\bibliographystyle{acl_natbib}

\clearpage
\appendix

\section{Data}

\subsection{Reasoning Type in EntailmentBank}
\label{app:EntailmentBank}
We list six different reasoning types in EntailmentBank dataset in Table~\ref{table_entb_types}. 

\subsection{Data Construction for DBpedia}
\label{app:DBpedia}
We constructed the DBpedia dataset to evaluate the compositional generalization of MORSE and baselines. Hence, DBpedia needs to contain several rules, and instances using one of these rules to process each step in multi-step reasoning. We extracted six reasoning rules as shown in Table~\ref{table_dbpedia_cons} from a rules pool. Following RuleBert \citep{saeed-etal-2021-rulebert} (Section 4.4 Chaining of Rule Executions), we generate hypotheses given existing rules over different relations and a depth D. Subsequently, we instantiate variables in rules and hypotheses from a name pool to generate instances. Rules and hypotheses are eventually transferred to natural language by pre-defined templates. 

\subsection{Data Construction for EntB-L and EntB-Sh}
\label{app:EntailmentBank_construct}
EntailmentBank contains 1,840 entailment trees showing how a hypothesis is entailed from a small number of relevant sentences. We constructed the EntailmentBank-Length (EntB-L) and EntailmentBank-Shape (EntB-Sh) for compositional generalization evaluation. In terms of EntB-L, we extracted data from the original dataset by the `length\_of\_proof' label. As for EntB-Sh, we extracted data from the original dataset by the `lisp\_proof' label. The shape of extracted trees are shown in Fig.\ \ref{fig_shape}.

\begin{table}[ht]
\resizebox{\columnwidth}{!}{
\begin{tabular}{l} \hline
Rules       \\ \hline
R1: child(B,A) $\rightarrow$ parent(A,B) \\ 
R2: child(A,C) $\wedge$ parent(C,B) $\rightarrow$ spouse(A,B)\\
R3: spouse(A,C) $\wedge$ parent(B,C) $\rightarrow$ negspouse(A,B)\\ 
R4: relative(A,C) $\wedge$ child(C,B) $\rightarrow$ relative(A,B)\\ 
R5: parent(A,C) $\wedge$ spouse(B,C) $\rightarrow$ parent(A,B) \\
R6: parent(A,C) $\wedge$ parent(B,C) $\rightarrow$ relative(A,B) \\ \hline
\end{tabular}}
\caption{Rules applied in DBpedia datasets.}
\label{table_dbpedia_cons}
\end{table}

\subsection{Data Statistic for Dataset}
\label{app:data_statics}
Table \ref{table_dataset} provides detailed data statistics of EntailmentBank and DBPedia. It contains the general data information for each dataset, and the data partitions we created and used for MORSE. 20\% of training data is adopted for validation. 

\begin{table}[t]
\resizebox{\linewidth}{!}{
\begin{tabular}{@{}lcc|lcccccc|lccc@{}} \hline
Dataset &EntB  &DBP& &\multicolumn{3}{c}{EntB-L(ength)} &\multicolumn{3}{c}{DBP-L(ength)}& &\multicolumn{3}{c}{EntB-Sh(apes)}\\
partitions        &      &   & & tr & ft & te & tr & ft& te & & tr & ft & te\\\hline
\#avg.nodes &7.6  &4   &L$_1$&430 &/ &/ &1800 &/ &/&A1 &390&/&/\\
\#avg.steps &3.2  &1.7 &L$_2$&450 &/ &/ &1800 &/ &/&A2&391&/&/\\
\#reas.types &6  &6&L$_3$&/ &300 &101 &/ &160 &391 &A3&219&/&/\\
\#examples &1840  &4560 & & & & & & &&B1 &/&79&36\\  
&&&&&& & & &&B2 &/&63&26\\
&&&&&& & & &&B3 &/&64&39\\ \hline
&&& all &880 &&&3600& && all &1000&206&101\\ \hline 
\end{tabular}}
\caption{Data statistics of Ent(ailment)B(ank) and DBP(edia). We split data into different partitions, including tr(ain), f(ine-)t(une) and te(st). L$_n$ denotes different lengths, and A*, B* means various shapes.}
\label{table_dataset}
\end{table}

\subsection{Data Statistic for Correlation Analysis}
\label{app:module_case}
To visualize the correlations between modules and rules, we constructed a new group of samples containing three reasoning steps. We select samples: i) contain three reasoning steps, ii) have correct predictions for the first two reasoning steps, but iii) the third step is incorrectly predicted in case a certain head is removed. The number of selected samples for each head is represented in Table~\ref{table_samples}. We then count samples in each head over different rules and show the correlations in Fig.\ \ref{fig_case}.

\begin{table}[h]
\resizebox{0.8\columnwidth}{!}{
\begin{tabular}{lcccccc} \hline
 &H1 &H2 &H3 &H4 &H5 &H6  \\ \hline
cases &126 &104 &137 &118 &104 &126\\ \hline
\end{tabular}}
\caption{Rules applied in DBpedia datasets.}
\label{table_samples}
\end{table}

\begin{table*}[ht]
\resizebox{\columnwidth}{!}{
\begin{tabular}{ll} \hline
 Reasonoing Types    & Example \\ \hline
\multirow{3}{*}{Substitution} &s1: when a light wave hits a reflective object, the light wave will be reflected  \\
                  &s2: a mirror is a kind of reflective object \\
                  &int: when a light wave hits a mirror, the light wave will be reflected \\ \hline
\multirow{3}{*}{Inference from Rule} &s1: puddles of water are outside during the day \\ 
                  &s2: if something is outside during the day then that something will receive sunlight  \\
                  &int: puddles of water will receive sunlight \\ \hline
\multirow{3}{*}{Further Specification or Conjuction} &s1: an animal requires warmth for survival as the season changes to winter \\
                  &s2: thick fur can be used for keeping warm \\
                  &int: thick fur can be used for keeping warm as the season changes to winter \\ \hline
\multirow{3}{*}{Infer Class from Properties} &s1: A compound is made of two or more elements chemically combined \\
                  &s2: sodium chloride is made of two elements chemically combined \\
                  &int: sodium chloride is a kind of compound \\ \hline
\multirow{3}{*}{Property Inheritance} &s1: an animal’s shell is usually hard \\
                  &s2: something hard can be used for protection \\
                  &int: an animal’s shell is usually hard for protection\\ \hline
\multirow{4}{*}{Sequential Inference} &s1: In molecular biology, translation follows transcription \\
                  &s2: transcription is when genetic information flows from DNA to RNA \\
                  &s3: translation is when genetic information flows from RNA to proteins \\
                  &int: In molecular biology, genetic information flows from DNA to RNA to proteins \\ \hline
\end{tabular}}
\caption{Six different reasoning types in EntailmentBank \citep{dalvi-etal-2021-explaining}}
\label{table_entb_types}
\end{table*}

\section{Experimental Details}
\label{app:parameters}
\textbf{MORSE} We conduct out-of-distribution experiments for increasing lengths and shapes of reasoning trees on two benchmarks, to test MORSE’s generalization abilities. Our experiments are run on Nvidia GTX 1080 Ti. As for length compositional test, MORSE (T5-Small and T5-Large) is trained for 33k steps and fine-tuning 4.5k steps on EntailmentBank-Length; trained for 8.1k steps and fine-tuning 0.6k steps on DBpedia-Length. In shape compositional test, MORSE is trained 25k steps and fine-tuning 5k steps. \par
\textbf{Baselines} Since ProofWriter-It and Entailment Writer are all T5-based baselines, we keep their settings as same as MORSE. In terms of Prover, we choose to use BERT-base-uncased version, given its parameters approach T5-small. We use the grid search technology for generation and select the best result. Its learning rate is 3e-5, trained for 36k steps and fine-tuning 4.5k steps on EntailmentBank-Length. In shape compositional test, Prover is trained 27k step and fine-tuning 5.5k steps. 

\begin{table*}[t]
\centering
\resizebox{\columnwidth}{!}{
\begin{tabular}{@{}lcccccc@{}} \hline
& \multicolumn{6}{c}{Original EntailmentBank Dataset (EntB-Orig)} \\ \cmidrule(r){2-7} 
\multirow{3}{*}{Models} & \multicolumn{2}{c}{Leaves} & \multicolumn{2}{c}{Steps} & \multicolumn{2}{c}{Intermediates} \\ \cmidrule(r){2-3} \cmidrule(r){4-5} \cmidrule(r){6-7}  
 &F1  &AllCorrect &F1  &AllCorrect  &F1  &AllCorrect   \\ \cmidrule{1-7}
Task 1 (no-distractor) - EntailmentWriter - T511b &99.0 &89.4 &51.5 &38.2 &71.2 &38.5 \\
Task 1 (no-distractor) - EntailmentWriter - T5Large &98.7 &86.2 &50.5 &37.7 &67.6 &36.2 \\
Task 1 (no-distractor) - MORSE (ours) - T5Large &98.09(0.24) &86,37(0.11) &51.11(0.84) &39.70(0.77) &69.79(0.09) & 40.97(0.34) \\  \cmidrule{1-7}
Task 1 (no-distractor) - EntailmentWriter - T5Small &98.40(0.41) &86.18(0.25) &41.72(0.96) &34.11(0.38) & 56.95(0.21) &40.41(0.49) \\
Task 1 (no-distractor) - MORSE (ours) - T5Small &98.30(0.37) &86.47(0.21) &42.35(0.66) &35.00(0.32) &57.76(0.11) &40.88(0.51) \\  \cmidrule{1-7} \hline
Task 2 (distractor) - EntailmentWriter - T511b &89.1 &48.8 &41.4 &27.7 &66.2 &31.5 \\
Task 2 (distractor) - EntailmentWriter - T5Large &84.3 &35.6 &35.5 &22.9 &61.8 &28.5 \\
Task 2 (distractor) - MORSE (ours) - T5Large &83.17(0.95) &34.41(0.59) &34,46(0.62) &21.96(0.60) &60.50(0.19) &28.24(0.37)  \\  \cmidrule{1-7}
\end{tabular}}
\caption{Comparative results for Entailment Writer vs. MORSE on original EntailmentBank dataset for Task 1 and Task 2 with different T5 model sizes}
\label{table_original_data_rst}
\end{table*}


\section{Comparative results on original EntailmentBank dataset}
\label{app:original_rst}
We conduct experiments of Task 1 and Task 2 from \citet{dalvi-etal-2021-explaining} on the \textit{original EntailmentBank dataset and splits}. The train, dev and test sets contain 1,313, 187 and 340 instances. Task 2 includes non-fitting distractor sentences in the input. We compare differently scaled T5 models to assess differences relating from T5 model sizes: T511b, T5large. EntailmentWriter (EW) is equivalent to MORSE modulo its modulated reasoning cell. For EW we show published results from \citet{dalvi-etal-2021-explaining}; for MORSE we report averaged results over three runs w/ standard deviation in brackets, for T5large. We observe comparable or superior results of MORSE w/T5large over EW w/t5large, especially for the difficult Steps (entailment tree structure) and Intermediates (inferred intermediate node label) evaluation criteria for Task 1. For Task 2, which poses a challenge by including noisy distractors, MORSE is still competitive, with ca. 1 percentage point distance. Comparing results of EW w/T511b vs. MORSE w/T5large shows that can MORSE rival and even outperform EW using T511b, for Steps and Intermediats Accuracies in Task 1, but not for the more difficult Task 2. The experiment shows that despite using a variation of the dataset in our main experiments to focus on MORSE's generalization abilities, it is still competitive on the original dataset and data distributions.

\section{Morse on powerful backbones}
\label{app:powerful_back}
\begin{table*}[t]
\centering
\resizebox{1\columnwidth}{!}{
\begin{tabular}{@{}lcccccccccccc@{}} \hline
& \multicolumn{6}{c}{EntailmentBank-Length (EntB-L)} & \multicolumn{6}{c}{DBpedia-Length (DBP-L)} \\ \cmidrule(r){2-7} \cmidrule(r){8-13} 
\multirow{3}{*}{Models} & \multicolumn{2}{c}{Leaves} & \multicolumn{2}{c}{Steps} & \multicolumn{2}{c}{Intermediates} & \multicolumn{2}{c}{Leaves} & \multicolumn{2}{c}{Steps} & \multicolumn{2}{c}{Intermediates}\\ \cmidrule(r){2-3} \cmidrule(r){4-5} \cmidrule(r){6-7} \cmidrule(r){8-9} \cmidrule(r){10-11} \cmidrule(r){12-13} 
 &F1  &AllCorrect &F1  &AllCorrect  &F1  &AllCorrect &F1  &AllCorrect &F1  &AllCorrect  &F1  &AllCorrect   \\ \cmidrule{1-13}
EntWriter (T5-Large) &99.78 &98.02 &52.80 &40.92 &56.62 &36.63 &99.36 &95.52 &82.49 &80.11 &88.98 &83.89 \\ 
MORSE (T5-Large) &\textbf{99.82}(+0.04) &\textbf{98.68}(+0.66) &\textbf{53.31}(+0.51) &\textbf{42.57}(+1.65) &\textbf{57.78}(+1.16) &\textbf{37.29}(+0.66) &\textbf{99.53}(+0.17) &\textbf{96.68}(+1.16) &\textbf{86.79}(+4.30) &\textbf{83.76}(+3.65) &\textbf{92.62}(+3.64) &\textbf{86.70}(+2.81) \\ \hline
EntWriter (Flan-T5-Large) &99.78 &98.02 &53.18 &41.58 &57.93 &39.13 &99.53 &95.52 &84.98 &83.12 &91.27 &84.14 \\
MORSE (Flan-T5-Large) &100.00(+0.22) &100.00(+1.98) &55.51(+2.33) &43.56(+1.98) &58.67(+0.74) &39.60(+0.47) &99.53(-0) &96.68(+1.16) &87.21(+2.23) &83.76(+0.64) &93.41(+2.14) &86.70(+2.56) \\  \cmidrule{1-13}\cmidrule{1-13} \hline
EntWriter-0-shot (T5-Large)  &97.06 &85.73 &18.44 &- &24.21 &- &90.09 &29.27 &16.94 &- &32.43 &-      \\
MORSE-0-shot (T5-Large) &\textbf{97.89}(+0.83) &\textbf{86.83}(+1.10) &\textbf{19.14}(+0.70) &- &\textbf{25.42}(+1.21) &- &\textbf{89.82}(-0.17) &\textbf{30.05}(+0.78) &\textbf{18.41}(+1.47) &- &\textbf{33.45}(+1.02) &- \\ \cmidrule{1-13}
EntWriter-0-shot (Flan-T5-Large) &98.79 &91.09 &20.59 &- &31.68 &- &90.05 &30.69 &18.46 &- &33.30 &-  \\
MORSE-0-shot (Flan-T5-Large) &99.82(+1.03) &92.31(+1.22) &21.22(+0.63) &- &32.07(+0.39) &- &91.96(+1.91) &31.28(+0.59) &21.99(+3.53) &- &33.92(+0.62) &-  \\  \cmidrule{1-13}
\end{tabular}}
\caption{Results on EntailmentBank-L(ength) and DBpedia-L(ength) for compositional generalization evaluation based on Flan-T5. (+num) indicates the improvement of MORSE compared to the strong baseline EntWriter.}
\label{table_flan}
\end{table*}

To further investigate the effectiveness of MORSE, we conduct experiments for MORSE and the most competitive baseline EntWriter on a more powerful backbone, e.g., Flan-T5 \citep{chung2022scaling}. Table \ref{table_flan} shows results. We find that: i) compared to T5, FLAN-T5 has generally better results for both models in both settings (fine-tuning and zero-shot). With FLAN-T5, our extension with MORSE still has superior results compared to the original T5 model. That is, our conclusions remain the same with this new backbone. ii) for both EntWriter and MORSE, FLAN-T5 shows increased performance in the zero-shot setting. This indicates that FLAN-T5 may serve as a better model variant to address zero-shot setting – which is expected for an instruction-tuned model.

\begin{figure}[t]
\centering
\includegraphics[scale=0.40,trim=3 0 0 0]{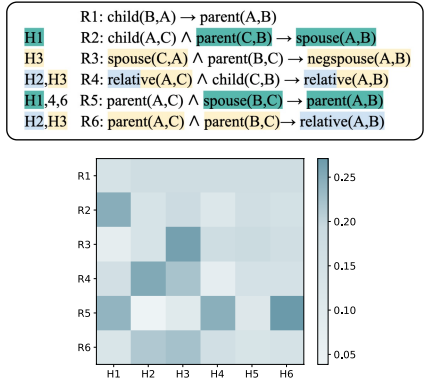}
\caption{Rules characteristics that pass through a specific head.}
\label{fig_cor_ana}
\end{figure}

\section{Characteristics in Correlation Analysis}
\label{app:corr_ana}
Figure \ref{fig_cor_ana} (from main text, reproduced for convenience) provides rules characteristics that pass through a specific head: 1) R5 and R2 are similar by changing `parent' to `child' between A and C; 2) R4 and R6 both use the predicate `relative' and share the same relation by changing `parent' to `child' between B and C; 3) R3, R4, and R6 share similar predicates ('spouse' in R3, 'relative' in R4 and 'parent' in R6). With these characteristics, we can roughly understand that each head is softly correlated with rules, sometimes distracted by similar rules.

\end{document}